\documentclass[sigconf]{acmart}

\AtBeginDocument{%
  }

\setcopyright{acmcopyright}
\copyrightyear{2023} 
\acmYear{2023} 
\setcopyright{acmlicensed}\acmConference[MM '23]{Proceedings of the 31st ACM International Conference on Multimedia}{October 29-November 3, 2023}{Ottawa, ON, Canada}
\acmBooktitle{Proceedings of the 31st ACM International Conference on Multimedia (MM '23), October 29-November 3, 2023, Ottawa, ON, Canada}
\acmPrice{15.00}
\acmDOI{10.1145/3581783.3611804}
\acmISBN{979-8-4007-0108-5/23/10}

\usepackage{subfigure}
\usepackage{bbding}
\usepackage{balance}
\usepackage{multirow}

\begin{document}

\title[SimulFlow: Simultaneously Extracting Feature and Identifying Target for UVOS]{SimulFlow: Simultaneously Extracting Feature and Identifying Target for Unsupervised Video Object Segmentation}


%
%

\author{Lingyi Hong\textsuperscript{\rm 1}, Wei Zhang\textsuperscript{\rm 1}*, Shuyong Gao\textsuperscript{\rm 1,}, Hong Lu\textsuperscript{\rm 1}*, WenQiang Zhang\textsuperscript{\rm 1,2}} 
\authornote{Corresponding authors}

\affiliation{%
 \institution{\textsuperscript{\rm 1}Shanghai Key Laboratory of Intelligent Information Processing, \\ School of Computer Science, Fudan University, Shanghai, China}
 \city{\textsuperscript{\rm 2} Engineering Research Center of AI \& Robotics, Ministry of Education, \\Academy for Engineering \& Technology, Fudan University, Shanghai, China\\}
 \country{honglyhly@gmail.com}
}

\renewcommand{\shortauthors}{Lingyi Hong et al.}

\begin{abstract}
  A clear and well-documented \LaTeX\ document is presented as an
  article formatted for publication by ACM in a conference proceedings
  or journal publication. Based on the ``acmart'' document class, this
  article presents and explains many of the common variations, as well
  as many of the formatting elements an author may use in the
  preparation of the documentation of their work.
\end{abstract}





\received{20 February 2007}
\received[revised]{12 March 2009}
\received[accepted]{5 June 2009}

\begin{abstract}
  Unsupervised video object segmentation (UVOS) aims at detecting the primary objects in a given video sequence without any human interposing. Most existing methods rely on two-stream architectures that separately encode the appearance and motion information before fusing them to identify the target and generate object masks. However, this pipeline is computationally expensive and can lead to suboptimal performance due to the difficulty of fusing the two modalities properly. In this paper, we propose a novel UVOS model called SimulFlow that simultaneously performs feature extraction and target identification, enabling efficient and effective unsupervised video object segmentation. Concretely, we design a novel SimulFlow Attention mechanism to bridege the image and motion by utilizing the flexibility of attention operation, where coarse masks predicted from fused feature at each stage are used to constrain the attention operation within the mask area and exclude the impact of noise.  Because of the bidirectional information flow between visual and optical flow features in SimulFlow Attention, no extra hand-designed fusing module is required and we only adopt a light decoder to obtain the final prediction. 
  We evaluate our method on several benchmark datasets and achieve state-of-the-art results. Our proposed approach not only outperforms existing methods but also addresses the computational complexity and fusion difficulties caused by two-stream architectures. Our models achieve 87.4\% $\mathcal{J} \& \mathcal{F}$ on DAVIS-16 with the highest speed (63.7 FPS on a 3090) and the lowest parameters (13.7 M). Our SimulFlow also obtains competitive results on video salient object detection datasets.
\end{abstract}

\maketitle

\begin{figure}[htbp]
  \centering
  \includegraphics[width=\linewidth]{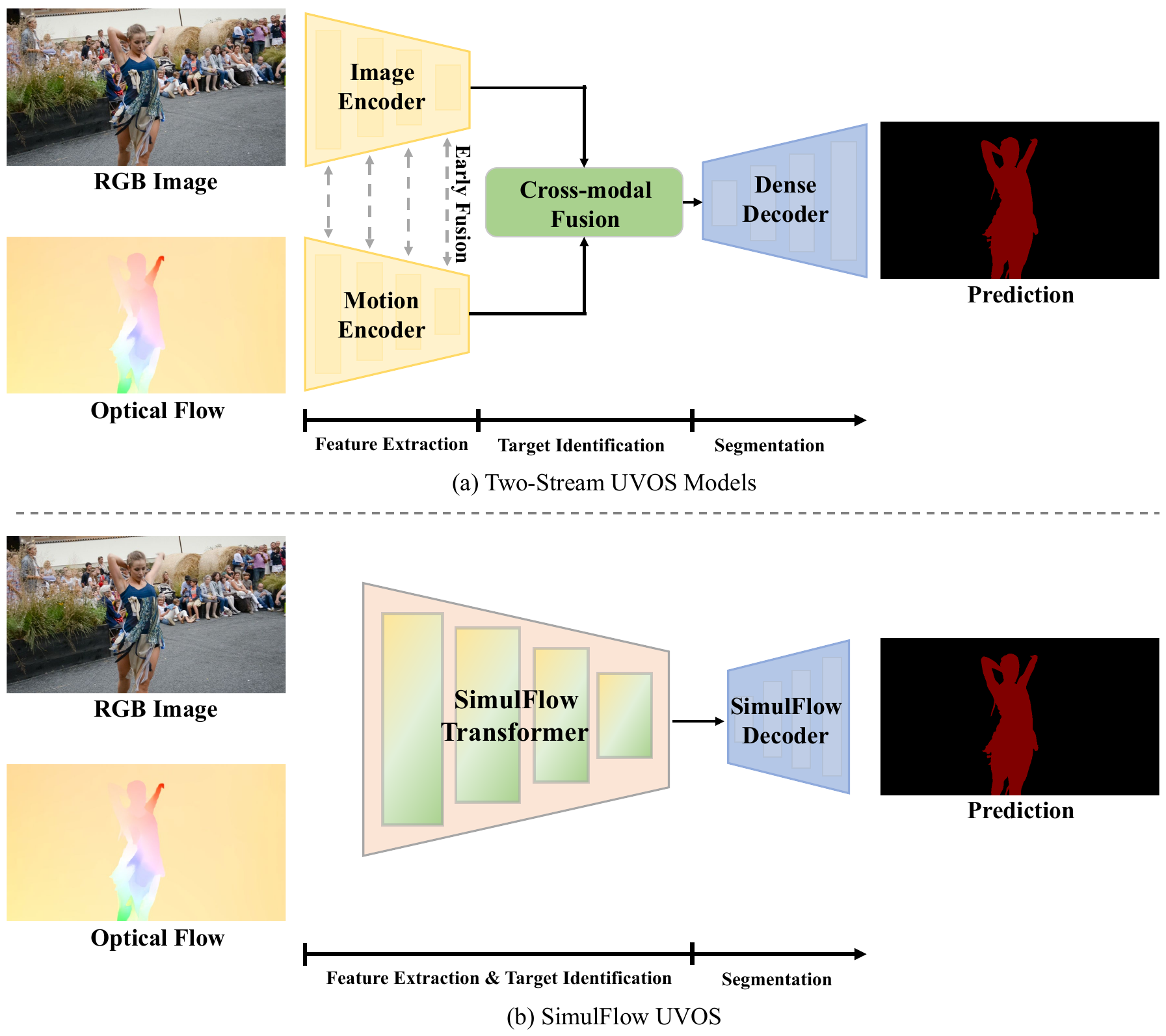}
  
  \caption{Comparison between dominant UVOS framework with our SimulFlow. (a) Previous separates feature extraction and cross-modal fusion. (b) SimulFlow consists of a SimulFlow Transformer encoder and a light decoder, which unifies the feature extraction and target identification jointly.}
  \label{fig:overview}
\end{figure}

\section{Introduction}
Unsupervised video object segmentation (UVOS) aims to automatically locate and segment the primary objects in each frame of a video without requiring any prior knowledge. UVOS has numerous practical applications, such as video compression~\cite{itti2004automatic}, autonomous driving~\cite{chen2015deepdriving,geiger2012we}, visual tracking~\cite{wu2014weighted}, and visual surveillance~\cite{xing2010multiple,tian2005robust}.

\begin{figure*}[htbp]
	\centering 
	%\subfigbottomskip=2pt 
	\subfigcapskip=-5pt 
	\subfigure{
		\includegraphics[width=0.42\linewidth]{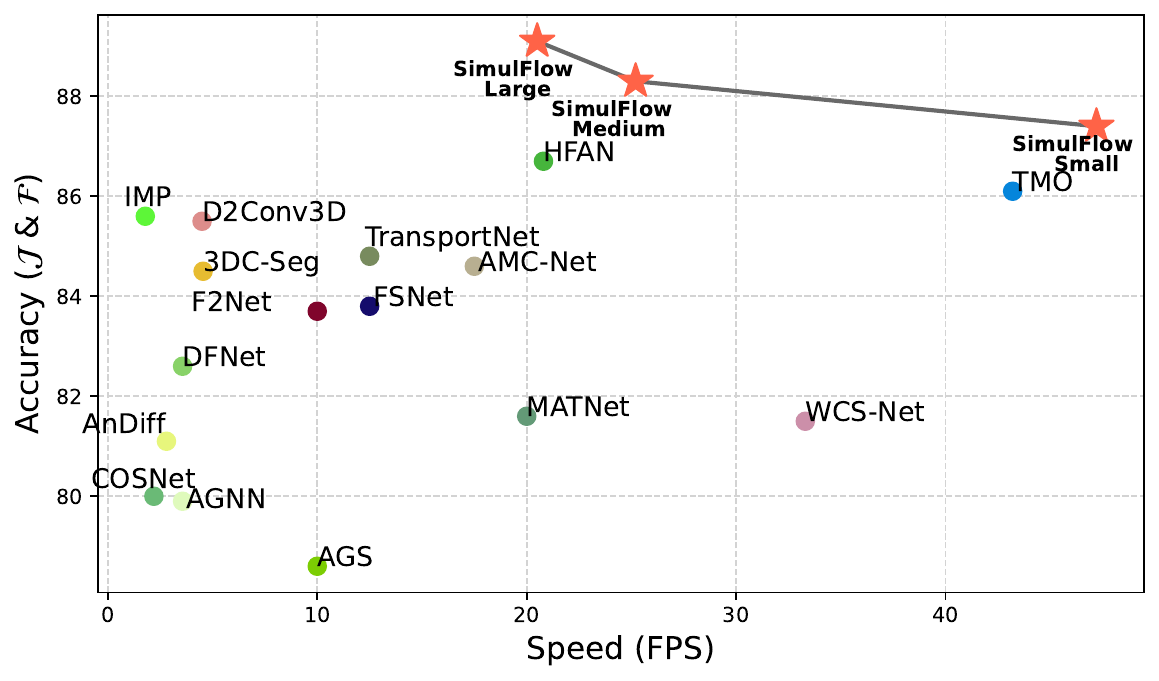}}
	\quad
	\quad
	\quad
	\subfigure{
		\includegraphics[width=0.42\linewidth]{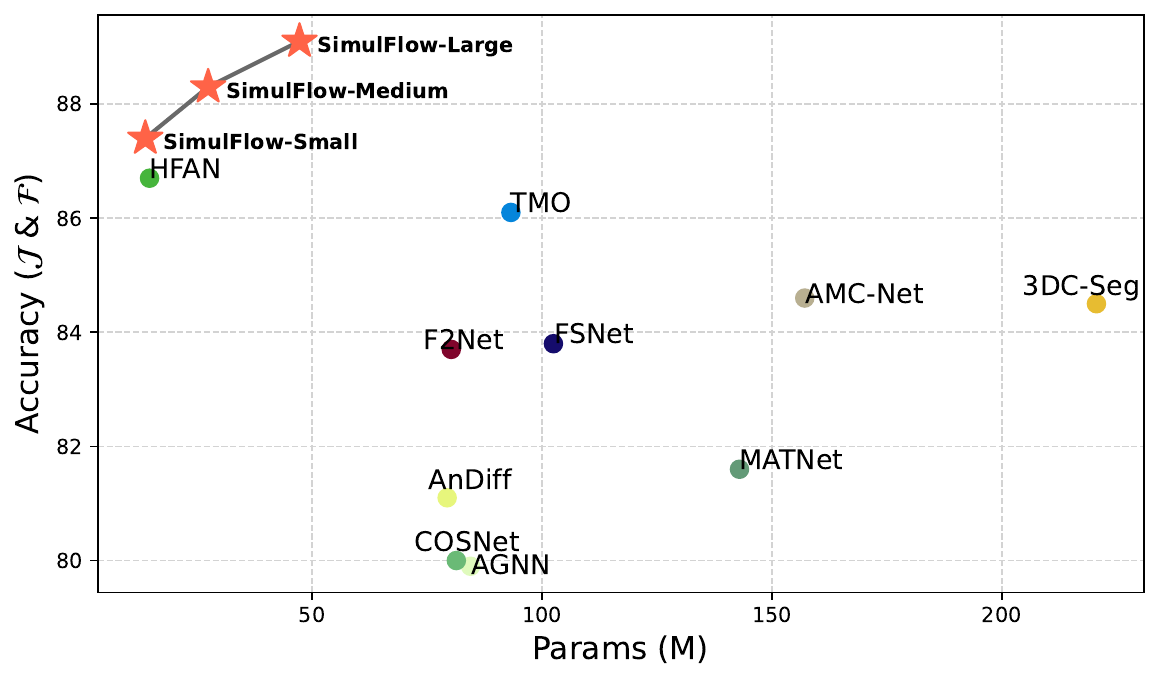}}
    \caption{Comparison with existing UVOS models on DAVIS-16. Our SimulFlow achieves balance between accuracy and efficiency.}
    \label{fig:per_compare}
\end{figure*}

In contrast to semi-supervised video object segmentation (SVOS) \cite{xu2018youtube,perazzi2016benchmark} where the groundtruth mask of the target object is provided in the first frame, UVOS faces the challenge of segmenting the primary object and distinguishing its pixels from the complex and diverse backgrounds without any manual guidance. To tackle this issue, a commonly-used approach is to utilize optical flow to capture motion information inspired by the observation that a salient object usually exhibits distinctive movements compared to the background. \cite{papazoglou2013fast,tokmakov2019learning} merely use motion information and may lose track if the object is moving slowly or is stationary. \cite{faisal2019exploiting,yang2021learning,zhang2021deep,ren2021reciprocal,zhou2020motion,pei2022hierarchical,cho2023treating} introduce appearance feature from RGB image in a two-stream structure to compensate for the shortage of motion descriptions on semantic representations as shown in Figure~\ref{fig:overview} (a). These models separate the pipeline into three components to accomplish the segmentation: feature extraction, target identification, and segmentation. In feature extraction part, two independent encoders are employed to extract the appearance feature and motion feature, respectively. It is worth noting that some methods, such as \cite{ren2021reciprocal,zhang2021deep,ji2021full}, may include early fusion between appearance and optical flow information during different stages of the encoder. After feature extraction, a cross-modal fusion module is designed to fuse the appearance and motion features to identify the target object, followed by a segmentation head to obtain the mask prediction. Although these two-stream pipelines which perform feature extraction and target identification sequentially have achieved significant advancements in UVOS, these models may suffer from dense computation and suboptimal performance due to the difficulty of properly fusing appearance and flow modalities. These algorithms have a number of limitations. Firstly, using two separate encoders to extract features makes the encoder unaware of the target object, which results in extracted features that are not focused on the specific object in each frame. The image encoder is pretrained for generic object recognition, which may not capture crucial structure information for video object segmentation. Additionally, the extracted motion feature may focus on all the objects in motion and introduce some noise. The limited target-background discriminative power makes extracted feature susceptible to continuously changing and complex scenes. Secondly, the two-stream structure is vulnerable to the performance-speed dilemma, which requires an extra cross-model fusion module for target localization. The complex fusion module brings about a large number of parameters and is difficult to optimize. 

In this paper, we propose a unified unsupervised video object segmentation framework called SimulFlow to address the above challenges of UVOS by simultaneously extracting features from both frames and optical flow fields, and fusing them to perform target identification. The advantages of SimulFlow can be summarized as follows. Firstly, the appearance feature is extracted dynamically and is more specific to the target object, which can contain richer information about the target. Secondly, the free information integration between appearance and motion information is allowed by utilizing our SimulFlow Attention. In contrast to the hand-designed fusion module which may not be able to integrate the two features well, our SimulFlow attention is built upon vanilla attention operation and is a very flexible architecture with dynamic and global modeling capacity to conduct efficient mutual integration. In a nut shell, because no extra cross-modal fusion module is required and the feature is stronger with a better perception of target objects, a light and simple decoder is sufficient to obtain the final mask prediction. As shown in Figure~\ref{fig:per_compare}, these advantages allow SimulFlow to achieve a balance between accuracy and efficiency.

\begin{figure*}[htbp]
  \centering
  \includegraphics[width=\linewidth]{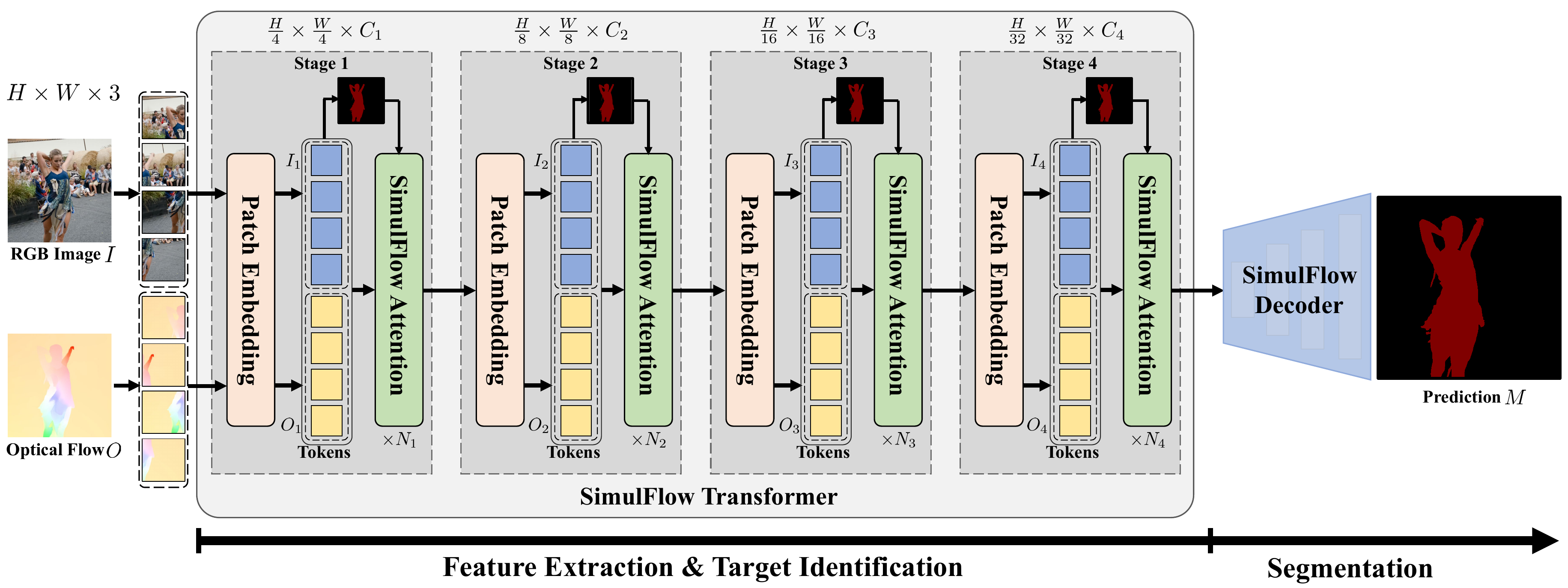}
  \caption{Overview of SimulFlow. Image and corresponding optical flow are fed into SimulFlow Transformer to perform simultaneous feature extraction and target identification. Then a light decoder is utilized to obtain final prediction.}
  \label{fig:model}
\end{figure*}

Concretely, we propose a novel SimulFlow Attention module to perform feature extraction and target identification simultaneously. It combines self-attention and cross-attention mechanisms, where self-attention extracts features independently from each modality, and cross-attention exchanges information between appearance and motion modalities to identify target. We also introduce a mask attention operation to reduce the negative impact of noise in optical flow. This mask attention operation generates a coarse mask based on the previous stage's output and then constrains the attention operation within the mask area to discard the background and exclude the influence of noise, thus boosting segmentation accuracy.

The main contributions of this paper are summarized as follows:
\begin{itemize}
  \item We propose a novel and simple unsupervised video object segmentation framework SimulFlow, which simultaneously extracts target-specific features and performs integration between appearance and motion information.
  \item We propose simulflow attention module which enables mutual guidance between appearance and motion features, while also excluding the negative impact of noise in optical flow through the insertion of a coarse mask.
  \item SimulFlow strikes an excellent balance between speed (63.7 FPS on a 3090 GPU) and performance ($87.4 \% \mathcal{J} \& \mathcal{F}$), surpassing previous models on multiple UVOS benchmarks with the fewest parameters and highest speed.
\end{itemize}

\begin{figure*}[htbp]
  \centering
  \includegraphics[width=\linewidth]{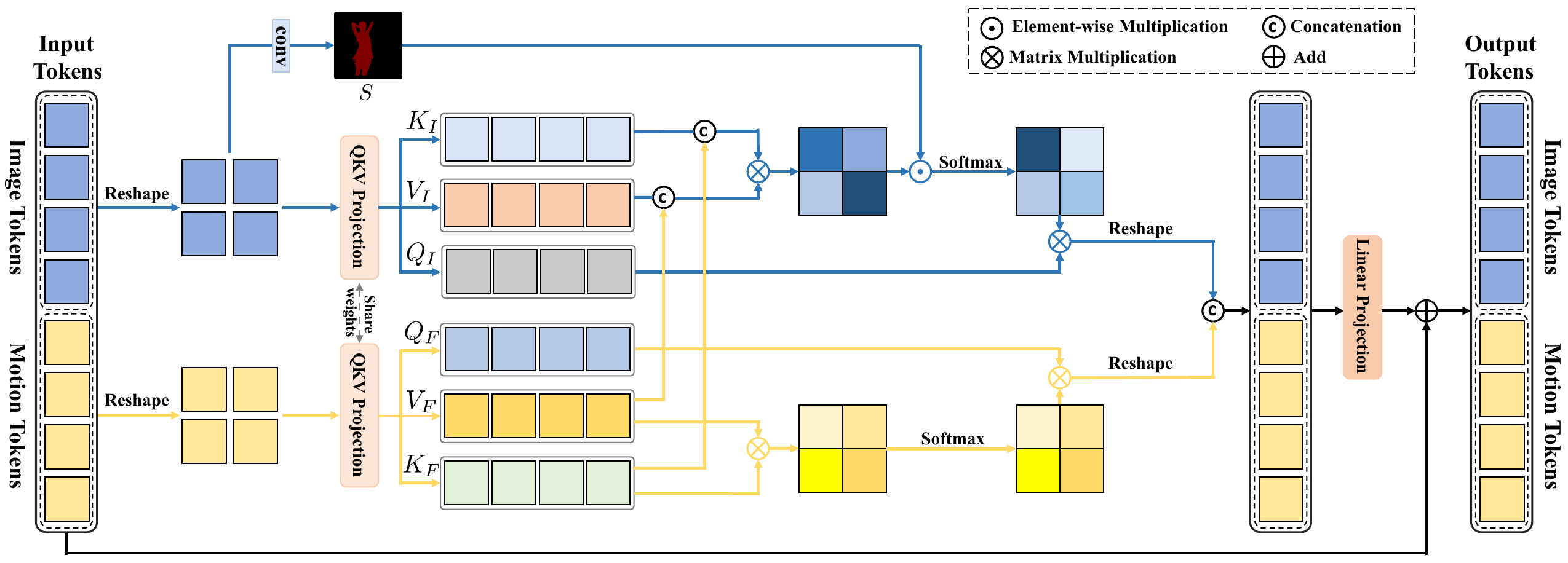}
  \caption{SimulFlow attention consists of self attention and cross attention. Self attention is responsible for extracting feature and cross attention allow the cross-modal integration. SimulFlow attention is the core of SimulFlow. }
  \label{fig:attn}
\end{figure*}

\section{Related Work}
\subsection{Semi-supervised Video Object Segmentation}
Semi-supervised video object segmentation (SVOS)~\cite{perazzi2016benchmark,xu2018youtube,hong2022lvos,ding2023mose} aims at highlighting the target object in a video given the spefic object mask at the first frame. ~\cite{caelles2017one,robinson2020learning, park2021learning, bhat2020learning} require finetuning pretrained networks at test time on the first frame, which can be time-consuming. Approaches such as~\cite{hu2018videomatch,chen2018blazingly,shin2017pixel,bao2018cnn} use the ground truth of the first frame to locate the target object in the current frame, while ~\cite{perazzi2017learning,oh2018fast,hu2017maskrnn,khoreva2017lucid} propagate predicted mask frame-to-frame based on already segmented previous frame. ~\cite{voigtlaender2019feelvos,yang2020collaborative,johnander2019generative} combine the first frame and previous frame and achieve better performance. Memory mechanisms to store all historical frames have been used in ~\cite{oh2019video,9665289,seong2020kernelized,9942927,cheng2021rethinking,seong2021hierarchical,yang2021associating}. Although SVOS obtain promising performance, the requirement of manual annotated mask in the first frame greatly limits its application in real-world scenarios.

\subsection{Unsupervised Video Object Segmentation}
Compared to SVOS, unsupervised video object segmentation (UVOS) does not require any annotations and aims to identify the salient object in a video. Popular UVOS models typically employ a two-stream structure that separates feature extraction and cross-modal fusion by leveraging appearance and motion cues. ~\cite{wang2019learning,lu2020video,wang2021survey} directly learns the high consistency of visual attention from previous images to segment salient object. Despite the high segmentation accuracy of existing UVOS models~\cite{zhou2020motion,ren2021reciprocal,ji2021full,yang2021learning,zhang2021deep,pei2022hierarchical,cho2023treating} the two-stream structure requires an extra cross-model fusion module after feature extraction to identify the target, which may be computationally dense and lead to suboptimal performance.
In this paper, we propose a novel UVOS pipeline SimulFlow that combines feature extraction and cross-modal integration together. SimulFlow allows for simultaneous feature extraction and target identification, which is much more efficient and effective than traditional two-stream models.

\subsection{One-stream Tracking}
Tracking is a fundamental task in computer vision, aiming to locate an object in a video based on the groundtruth box in the first frame. Siamese networks are widely used in tracking algorithms, in which a siamese encoder extracts image features from the target and template frames separately, similar to UVOS. Subsequently, an integration module is utilized to model the relationship between the target and template frames to search for the target in the subsequent frames. Compared with two-stream tracking methods, MixFormer~\cite{cui2022mixformer}, OSTrack~\cite{ye2022joint}, and SimTrack~\cite{chen2022backbone} unify the feature extraction and information integration in a visual backbone. A Mixed Attention Module (MAM) is proposed in MixFormer to enable extensive communication during the feature extraction process. OSTrack and SimTrack utilize vanilla ViT~\cite{dosovitskiy2020image} as the backbone to locate target object.  One-stream tracking models  take RGB images as input and perform the target location by conducting matching between the two images. As opposed to one-stream tracking pipelines that predict the bounding box of the target in RGB images, our SimulFlow is specifically tailored to unsupervised video segmentation task, taking RGB image and optical flow as input and generating mask predictions. Moreover, our SimulFlow achieves the target identification through the mutual cooperation between appearance and motion information.

\section{Methodology}
In this section, we will describe our novel UVOS framework SimulFlow. RGB image and optical flow are taken as input into SimulFlow transformer for simultaneous feature extraction and target identification. After that, a light decoder is employed to obtain the final prediction. An overview of SimulFlow is shown in Figure~\ref{fig:model}. 

\subsection{SimulFlow}
To combine the feature extraction and mutual information flow, we build an end-to-end UVOS pipeline SimulFlow, which comprises two components: a SimulFlow transformer to identify the target and a lightweight SimulFlow decoder to generate mask predictions.

\textbf{SimulFlow Transformer.} 
SimulFlow transformer comprises four progressive stages, each of which has a patch embedding layer and $N_{i}$ SimulFlow attention layers, where $i$ denotes the index of the stage. Specifically, given the input RGB image $I$  with the size of $H \times W \times 3$ and corresponding optical flow $O$, we first split them into patches of size $4 \times 4$ and utilize patch embedding ;ayer to project the patches into feature space. Then, these patches are fed into a SimulFlow transformer with a hierarchical design to obtain multi-level feature representation of appearance and motion features. Appearance and motion feature tokens of the $i$-th stage ($i \in \{1,2,3,4\}$) are denoted as $I_{i} \in \mathbb{R}^{H_{i}W_{i} \times C_{i}}$ and $O_{i} \in \mathbb{R}^{H_{i}W_{i} \times C_{i}}$, respectively. $H_{i}W_{i}$ is the size of $I_{i}$ and $O_{i}$, and $C_{i}$ is the channel number of $I_{i}$ and $O_{i}$. $H_{i}W_{i}$ is set to $\frac{H}{2^{i+1}} \times \frac{W}{2^{i+1}}$. Because of the simulflow attention module, feature extraction and target identification can be performed simultaneously, and the extracted features of each stage are specific to the target object after cross-model integration. We will detail the structure of simulflow attention module in Section~\ref{sec:attn}.

\begin{table*}[htbp]
	\centering
	\caption{Quantitative evaluation on the DAVIS-16~\cite{perazzi2016benchmark}, FBMS~\cite{ochs2013segmentation}, and YouTube-Objects~\cite{prest2012learning}. OF and PP indicate the use of optical flow estimation models and post-processing techniques, respectively. $\#$Params is measured in MB. $\dag$ denotes the speed on a single NVIDIA RTX 2080Ti GPU, and * denotes the speed on a single NVIDIA GTX 3090 GPU.}
	\scriptsize
    \setlength{\tabcolsep}{3.7mm}
	\begin{tabular}{l c c c c c c ccc c c}
		\toprule
		\multirow{2}{*}{Method}  & \multirow{2}{*}{Publication} & \multirow{2}{*}{Resolution} & \multirow{2}{*}{OF} & \multirow{2}{*}{PP} & \multirow{2}{*}{$\#$Params} & \multirow{2}{*}{FPS} &\multicolumn{3}{c}{DAVIS-16} &\multicolumn{1}{c}{FBMS} &\multicolumn{1}{c}{YTO}\\
		\cline{8-12}
		 & & &  &  &   & & $\mathcal{G}_\mathcal{M}\uparrow$ & $\mathcal{J}_\mathcal{M}\uparrow$ & $\mathcal{F}_\mathcal{M}\uparrow$ & $\mathcal{J}_\mathcal{M}\uparrow$ & $\mathcal{J}_\mathcal{M}\uparrow$\\
		\midrule
        PDB~\cite{song2018pyramid} &ECCV'18 & $473\times 473$ & &\checkmark & - &20.0 &75.9 &77.2 &74.5 &74.0 & 65.5\\
		MOTAdapt~\cite{siam2019video} &ICRA'19 & - & &\checkmark & - & - &77.3 &77.2 &77.4 &- & 58.1\\
		AGS~\cite{wang2019learning} &CVPR'19 & $473\times 473$ & &\checkmark & - &10.0 &78.6 &79.7 &77.4 &- & 69.7\\
		COSNet~\cite{lu2019see} &CVPR'19 & $473\times 473$ & &\checkmark & 81.4 &- &80.0 &80.5 &79.4 &75.6 & 70.5\\
		AD-Net~\cite{yang2019anchor} &ICCV'19 & $480\times 854$ & &\checkmark  & 79.4 &4.00 &81.1 &81.7 &80.5 & - & -\\
		AGNN~\cite{wang2019zero} &ICCV'19 & $473\times 473$  & &\checkmark  & 82.4 &3.57 &79.9 &80.7 &79.1 &- & 70.8\\
		MATNet~\cite{zhou2020motion} & AAAI'20 & $473\times 473$   &\checkmark &\checkmark & 142.9 &20.0 &81.6 &82.4 &80.7 &76.1 & 69.0\\
		WCS-Net~\cite{zhang2020unsupervised} & ECCV'20 & $320\times 320$ & & & - & 33.3 &81.5 &82.2 &80.7 &- & 70.5\\
		DFNet~\cite{zhen2020learning} & ECCV'20 & - & &\checkmark & 64.7 &3.57 &82.6 &83.4 &81.8 &- & -\\
        3DCSeg~\cite{mahadevan2020making} & BMVC'20 & $480\times 854$  & & & 220.6 &4.55 &84.2 &84.2 &84.3 &- & -\\
		F2Net~\cite{liu2021f2net} & AAAI'21 & $473\times 473$ & &  & 80.3 &10.0 &83.7 &83.1 &84.4 &77.5 & -\\
		RTNet~\cite{ren2021reciprocal} & CVPR'21 & $384\times 672$ &\checkmark &\checkmark & - &- &85.2 &85.6 &84.7 &- & 71.0\\
		FSNet~\cite{ji2021full} & ICCV'21 & $352\times 352$ &\checkmark &\checkmark & 102.5 &12.5 &83.3 &83.4 &83.1 &-&-\\
		TransportNet~\cite{zhang2021deep} & ICCV'21 & $512\times 512$ &\checkmark & & - &12.5 &84.8 &84.5 &85.0 &78.7 & -\\
		AMC-Net~\cite{yang2021learning} & ICCV'21 & $384\times 384$ &\checkmark &\checkmark & 157.2 &17.5 &84.6 &84.5 &84.6 &76.5 & 71.1\\
		IMP~\cite{lee2022iteratively} & AAAI'22 & - & & & - &1.79 &85.6 &84.5 &86.7 &77.5 & -\\
		HFAN~\cite{pei2022hierarchical} & ECCV'22 & $512\times 512$ &\checkmark &  & 14.6 & 20.8 & 86.7 & 86.2 & 87.1 &- & 73.4\\
		PMN~\cite{lee2023unsupervised} & WACV'23 & - &  \checkmark & & - & - &85.9 &85.4 &86.4 &77.7 & 71.8\\
		TMO~\cite{cho2023treating} & WACV'23 & $384\times 384$ &\checkmark &  & 93.2 &43.2 &86.1 &85.6 &86.6 &79.9 & 71.5 \\
		\midrule
        SimulFlow-Small & ACM MM'23 & $384\times 384$  &\checkmark &  & \textbf{13.7} & \textbf{53.2}$\dag$/\textbf{81.1*} & 86.7 & 86.3 & 87.4 & 80.1 & 71.7 \\
        SimulFlow-Small & ACM MM'23 & $512\times 512$  &\checkmark &  & \textbf{13.7} & \textbf{47.2}$\dag$/\textbf{63.7*} & 87.4 & 86.9 & 88.0 & 80.4 & 72.9 \\
        SimulFlow-Medium & ACM MM'23 & $512\times 512$ &\checkmark &  & 27.4 & 25.2$\dag$/36.5* & 88.3 & 87.1 & 89.5 & 84.1 & 74.6 \\
        SimulFlow-Large & ACM MM'23 & $512\times 512$  &\checkmark &  & 47.2 & 20.5$\dag$/26.3* & \textbf{89.1} & \textbf{87.7} & \textbf{90.5} & \textbf{85.5} & \textbf{75.7} \\
		
		\bottomrule
\end{tabular} 

	\label{Table:results}
\end{table*}

\textbf{SimulFlow Decoder.} 
After extracting the appearance and motion features, we employ a light and simple decoder to obtain the final prediction. The decoder is similar to the one in~\cite{xie2021segformer}, which consists of four steps. In the first step, multi-level features $I_{i}$ and $O_{i}$ are reshaped to the size of $H_{i} \times W_{i} \times C$ to restore 2D shape and then fed into a MLP layer to unify channel dimension. In the second step, all these features are upsampled to size $\frac{H}{4} \times \frac{W}{4}$ using an upsampling function. In the third step, the upsampled features $F_i$ are concatenated together. Finally, a MLP layer is adopted to fuse the concatenated features.
%and another MLP layer is used to generate mask prediction $M'$ with a size of $\frac{H}{4} \times \frac{W}{4}$. 
The procedure of the decoder can be mathematically expressed as follows:
\begin{equation}
    \begin{aligned}
      F_{i}&=Linear(C_{i},C)(I_{i}+O_{i}),\forall i \\
      F_{i}&=Upsample(\frac{H}{4} \times \frac{W}{4})(F_{i}),\forall i \\
      F&=Linear(4C,C)(Concat(F_{1},F_{2},F_{3},F_{4})),
    \end{aligned}
\end{equation}

where $Linear(C_{in},C_{out})(\dots)$ denotes a linear projection and $C_{in}$ and $C_{out}$ are dimensions of input and output vector, respectively, $\text{Upsample}(S)$ denotes the upsampling function to the size of $S$, and $\text{Concat}(F_{1},F_{2},F_{3},F_{4})$ denotes the concatenation of the upsampled features $F_i$ from all the stages. After that, we adopt another MLP layer to generate mask prediction   $M'$ with a size of $\frac{H}{4} \times \frac{W}{4}$. Then we upsample $M'$ to the origin shape $H \times W$. Softmax and argmax operations are applied to $M'$ to obtain the final binary mask $M$, which can be phrased as:
\begin{equation}
    \begin{aligned}
        M'&=Linear(C,2)(F),\\
        M'&=Upsample(H \times W)(M'),\\
        M&=argmax(softmax(M')) \in \{0,1\}^{H \times W}.
    \end{aligned}
\end{equation}

\subsection{SimulFlow Attention}
\label{sec:attn}
SimulFlow attention is a key component of SimulFlow that allows for simultaneous feature extraction and target identification. The attention operation is conducted between two token sequences, i.e. image tokens and motion tokens, using an asymmetric design, as shown in Figure~\ref{sec:attn}. 

Specifically, inspired by ~\cite{cui2022mixformer}, for the given input token sequences, we first split them into two parts: image tokens and motion tokens, and reshape them into 2D feature maps. Then a linear projection layer is used to map each token sequence into its corresponding key, value, and query feature. It is worth noting that there may be a convolution layer to downsample the key and value matrices for efficiency. To perform the attention operation, all these features are flattened. The query, key, value features for image tokens are denoted as $Q_{I}$, $K_{I}$, $V_{I}$, and those for motion tokens are denoted as $Q_{O}$, $K_{O}$, $V_{O}$.  The fusion of image and motion features is performed in an asymmetric manner. Finally, the output tokens are concatenated and then fed into a linear projection layer.

We first denote the common attention operation as:

\begin{equation}
    \begin{aligned}
      A=Attn(Q,K,V)=Softmax(X)V=Softmax(\frac{QK^{\mathrm{T}}}{\sqrt{d}})V,\\
    \end{aligned}
\label{eq:co_attn}
\end{equation}
where $Q, K, V$ are the query, key, and value embeddings, $X$ is the correlation map between $Q$ and $K$, and $d$ is the dimension of key feature $K$. The asymmetric SimulFlow attention is defined as:
\begin{equation}
    \begin{aligned}
      K_{U}=&[K_{I}; K_{O}], V_{U}=[V_{I}; V_{O}],\\
      A_{I}&=Attn(Q_{I},K_{U},V_{U}),\\
      A_{O}&=Attn(Q_{O},K_{O},V_{O}),
    \end{aligned}
\label{eq:attn}
\end{equation}
where $A_{I}$ and $A_{O}$ are attention map of image and motion. The SimulFlow attention contains not only self attention but also cross attention. Self attention is responsible for feature extraction and cross attention enables the mutual information flow. It is worth noting that we disable the cross attention from image to motion to reduce the computational cost, because the information from appearance to motion may result in a negative influence, which will be elaborated in Section~\ref{sec:ablation}. Through the simple but efficient attention mechanism, we achieve jointly feature extraction and target identification. We further analyze the attention operation to explain why SimulFlow attention can unify these two parts from the perspective of attention mechanism. Equation~\ref{eq:attn} can be expanded to:

\begin{equation}
    \begin{aligned}
      A_{I}&=Softmax(\frac{Q_{I}K_{U}^{\mathrm{T}}}{\sqrt{d}})V_{U} \\
      &=Softmax(\frac{Q_{I}[K_{I}; K_{O}]^{\mathrm{T}}}{\sqrt{d}})[V_{I};V_{O}]\\
       &\triangleq [W_{II}, W_{IO}][V_{I}; V_{O}]\\
       &=W_{II}V_{I}+W_{IO}V_{O},
    \end{aligned}
\end{equation}
where $W_{IO}$ is a measure of the similarity between appearance and motion, and $W_{II}$ is similarity between appearance and itself. $W_{IO}V_{O}$ is responsible for the cross-model fusion and $W_{II}V_{I}$ extracts further appearance feature. Thus, our SimulFlow attention succeeds in combining feature extraction and target identification in a simple operation.

Additionally, to exclude the impact of distracting noise of optical flow, we further insert a coarse mask into the SimulFlow attention. For appearance features $I_{i}$ from $i$-th stage of the encoder, a convolution layer is implemented to inject it into a coarse mask prediction $S_{i} \in \mathbb{R}^{H_{i}\times W_{i}}$. We utilize $S_{i}$ to modulate attention matrix via:

\begin{equation}
    \begin{aligned}
      A_{I}=Softmax(X_{I}\odot Sigmoid(S_{i}))V_{U}, 
    \end{aligned}
\end{equation}
where $\odot$ denotes the element-wise multiplication. Through the soft mask restraint, we constrain the attention calculation within the mask area and decrease the negative influence of noise.

\begin{table*}
	\centering 
	\caption{Quantitative evaluation on the DAVIS-16~\cite{perazzi2016benchmark}, ViSal~\cite{li2013video}, and DAVSOD~\cite{fan2019shifting} for VSOD.}
    \scriptsize
    \setlength{\tabcolsep}{2.4mm}
	\begin{tabular}{cc|cccccccccccccc|c}
			\toprule
			\multicolumn{2}{c|}{\multirow{2}{*}{Metrics}}& SCNN & SCOM & DLVS & FGR & MBNM & PDB & MGA & RCR & SSAV  & PCSA  & WSV  & TENET & STVS  & DCF & SimulFlow   \\
			&&\cite{tang2018weakly} &\cite{chen2018scom} &\cite{wang2017video} &\cite{li2018flow} &\cite{li2018unsupervised} &\cite{song2018pyramid} &\cite{li2019motion} &\cite{yan2019semi} &\cite{fan2019shifting}  &\cite{gu2020pyramid}  &\cite{zhao2021weakly}  &\cite{ren2020tenet} &\cite{xie2021efficient}  &\cite{zhang2021dynamic} & Small   \\  \hline
			\multicolumn{1}{c}{\multirow{3}{*}{\textbf{DAVIS}}}
			& $MAE\downarrow$ & 0.064 & 0.048 & 0.055 & 0.043 & 0.031  & 0.028 & 0.022   & 0.027 & 0.028 & 0.023 & 0.037 & 0.021  & 0.023 & 0.016  & \textbf{0.009} \\
			& $S_{m}\uparrow$ & 0.783 & 0.832 & 0.802 & 0.838 & 0.887  & 0.882 & 0.910   & 0.886 & 0.893 & 0.902 & 0.828 & 0.905  & 0.892 & 0.914  & \textbf{0.937} \\
			& $F_{\beta}\uparrow$ & 0.714 & 0.783 & 0.721 & 0.783 & 0.862  & 0.855 & 0.892   & 0.848 & 0.861 & 0.88  & 0.779 & 0.894  & 0.865 & 0.900  & \textbf{0.936} \\ \hline
			\multicolumn{1}{c}{\multirow{3}{*}{\textbf{ViSal}}}
			& $MAE\downarrow$ & 0.071 & 0.122 & 0.048 & 0.045 & 0.020  & 0.032 & 0.017   & 0.027 & 0.020 & 0.017 & 0.041 & 0.021  & 0.013 & \textbf{0.010}  & 0.012 \\
			& $S_{m}\uparrow$ & 0.847 & 0.762 & 0.881 & 0.861 & 0.898  & 0.907 & 0.940   & 0.922 & 0.943 & 0.946 & 0.857 & 0.943  & \textbf{0.954} & 0.952  & 0.946 \\
			& $F_{\beta}\uparrow$ & 0.831 & 0.831 & 0.852 & 0.848 & 0.883  & 0.888 & 0.936   & 0.906 & 0.939 & 0.94  & 0.831 & 0.947  & 0.953 & \textbf{0.953}  & 0.943 \\ \hline
			\multicolumn{1}{c}{\multirow{3}{*}{\textbf{DAVSOD}}}
			& $MAE\downarrow$ & 0.127 & 0.219 & 0.129 & 0.095 & 0.109  & 0.116 & 0.083   & 0.087 & 0.092 & 0.086 & 0.103 & 0.078  & 0.086 & 0.074  & \textbf{0.069} \\
			& $S_{m}\uparrow$ & 0.680 & 0.603 & 0.664 & 0.701 & 0.646  & 0.698 & 0.741   & 0.741 & 0.724 & 0.741 & 0.705 & 0.753  & 0.744 & 0.741  & \textbf{0.771} \\
			& $F_{\beta}\uparrow$ & 0.541 & 0.473 & 0.541 & 0.589 & 0.506  & 0.572 & 0.643   & 0.653 & 0.603 & 0.655 & 0.605 & 0.648  & 0.650 & 0.660  & \textbf{0.722} \\ \hline

			\bottomrule
\end{tabular} 
	\label{tab:results_vsod}
\end{table*}

\textbf{Discussion.} One-stream tracking methods~\cite{cui2022mixformer,ye2022joint,chen2022backbone} also employ attention mechanisms similar to our SimulFlow. However, the purpose of the attention mechanism in these trackers is to model the matching relationship between the template and search area to locate the target object, whereas our SimulFlow attention is designed to perform cross-model integration between appearance and motion features to identify the salient object. While the key component of attention in one-stream tracking methods is matching, SimulFlow attention mechanism focuses on the synergy of two modal features to achieve salient object localization. Additionally, SimulFlow includes a mask constraint to exclude the noise of optical flow. Notably, SimulFlow is specifically designed for the unsupervised video object segmentation task, which is more challenging than tracking due to the requirement of pixel-wise classification.

\subsection{Training and Inference}

\textbf{Training.} 
Following previous works\cite{ren2021reciprocal,liu2021f2net}, we divide the training process into three stages. Firstly, we pretrain our model on the salient object detection dataset DUTS~\cite{wang2017learning} to prevent overfitting and facilitate model robustness. Due to the lack of optical flow in DUTS, we just pretrain SimulFlow without the cross-modal fusion. Then, we train the whole model with cross-model matching on YouTube-VOS~\cite{xu2018youtube}. Because there may exist more than one object in videos from YouTube-VOS, we regard them as a single object to obtain a binary mask. Optical flow map is generated by a popular optical flow estimation model RAFT~\cite{teed2020raft}. Finally, we finetune SimulFlow on DAVIS-16~\cite{perazzi2016benchmark}. We optimize SimulFlow by minimizing the cross entropy loss between mask prediction and groundtruth. Moreover, we also supervise the coarse mask at each stage for stable training. The overall loss function can be written as:

\begin{equation}
      L=L_{ce}(M,M_{G})+\lambda\sum_{i=1}^{4} L_{bce}(S_{i},M_{G}), 
\end{equation}
where $M_{G}$ is groundtruth, $L_{ce}$ and $L_{bce}$ refer to cross entropy and binary cross entropy loss, respectively. $\lambda$ is regularization parameter.

\textbf{Inference}
During inference, we resize each frame to 512$\times$512 without any data augmentation or human interaction and feed image and optical flow into SimulFlow. No post-processing techniques such as Conditional Random Field (CRF)~\cite{krahenbuhl2011efficient} are employed. The optical flow map is pre-computed and stored on disk. The time taken for optical flow estimation is excluded from the final time statistics following previous works~\cite{cho2023treating,lee2023unsupervised,pei2022hierarchical,yang2021learning,zhang2021deep,ji2021full,ren2021reciprocal}. All inference processes are performed on a single NVIDIA GTX 3090 GPU. For fair comparision, we also report the inference time on a NVIDIA RTX 2080Ti GPU.

\section{Experiments}
\label{sec:experiment}

\begin{table*}
    \scriptsize
    \caption{Ablation study on each component of SimulFlow on DAVIS-16~\cite{perazzi2016benchmark}. And the speed is calculated on a single 3090 GPU.}
    \label{tab:ablation}
    \centering
    \subtable[The effect of backbone.]{
        \renewcommand{\arraystretch}{1.76}
        \begin{tabular}{l l c c c}
    \toprule
    Model & Backbone & $\#$Params & FPS & $\mathcal{G}_\mathcal{M}$ \\
    \midrule
    SimulFlow-small & MiT-b1 & \textbf{13.7} & \textbf{63.7} & 87.4\\
    SimulFlow-medium & MiT-b2 & 27.4 & 36.5 & 88.3 \\
    SimulFlow-large & MiT-b3 & 47.2 & 26.3 & \textbf{89.1} \\
    \bottomrule

\end{tabular}

    \label{tab:ablation_baseline}
    }
    \subtable[The effect of cross attention.]{
        \renewcommand{\arraystretch}{1.41}
        \begin{tabular}{l c c}
    \toprule
    Variant & FPS & $\mathcal{G}_\mathcal{M}$ \\
    \midrule
    Baseline & \textbf{74.6} & 83.8\\
    Baseline + Motion to Image  & 63.7 & 85.6 \\
    Baseline + Image to Motion & 63.7 & 84.9 \\
    Baseline + Image \& Motion & 52.4 & \textbf{85.8} \\
    \bottomrule

\end{tabular}

        \label{tab:ablation_cross}
    }

    \qquad
    \subtable[The effect of mask operation.]{
        \renewcommand{\arraystretch}{1.17}
        \begin{tabular}{l c c c}
    \toprule
    Variant & $\#$Params & FPS & $\mathcal{G}_\mathcal{M}$ \\
    \midrule
    Baseline & \textbf{13.5} & \textbf{74.6} & 83.8 \\
    Baseline + Image Mask & 13.7 & 73.2 & \textbf{85.1} \\
    Baseline + Image Hard Mask & 13.7 & 73.2 & 84.7 \\
    Baseline + Motion Mask & 13.7 & 73.2 & 84.3 \\
    Baseline + Image \& Motion Mask & 13.7 & 72.6 & 84.4 \\
    \bottomrule

\end{tabular}

        \label{tab:ablation_mask}
    }
    \subtable[The effect of simulflow attention.]{
        \begin{tabular}{c c c c c c}
    \toprule
    \multicolumn{4}{c}{Stage} & \multirow{2}{*}{FPS} & \multirow{2}{*}{$\mathcal{G}_\mathcal{M}$} \\
    \cline{0-3}
    1 & 2 & 3 & 4 &  &  \\
    \midrule
     &  &  &  & \textbf{74.6} & 83.8 \\
    \checkmark &  &  &  & 70.3 & 84.7 \\
    \checkmark & \checkmark &  &  & 68.4 & 85.5 \\
    \checkmark & \checkmark & \checkmark &  & 65.1 & 86.6 \\
    \checkmark & \checkmark & \checkmark & \checkmark & 63.7 & \textbf{87.4} \\
    \bottomrule

\end{tabular}

        \label{tab:ablation_stage}
    }
    \subtable[The effect of training stratergy.]{
        \renewcommand{\arraystretch}{1.41}
        \begin{tabular}{c c c c}
    \toprule
    DUTS & YTB & DAVIS & $\mathcal{G}_\mathcal{M}$ \\
    \midrule
     &   & \checkmark & 80.2 \\
      & \checkmark  & \checkmark & 82.4 \\
     \checkmark  &  & \checkmark & 83.5 \\
     \checkmark  & \checkmark & \checkmark & \textbf{87.4} \\
    
    \bottomrule

\end{tabular}

        \label{tab:ablation_train}
    }
\end{table*}

\begin{figure*}[htbp]
  \centering
  \includegraphics[width=\linewidth]{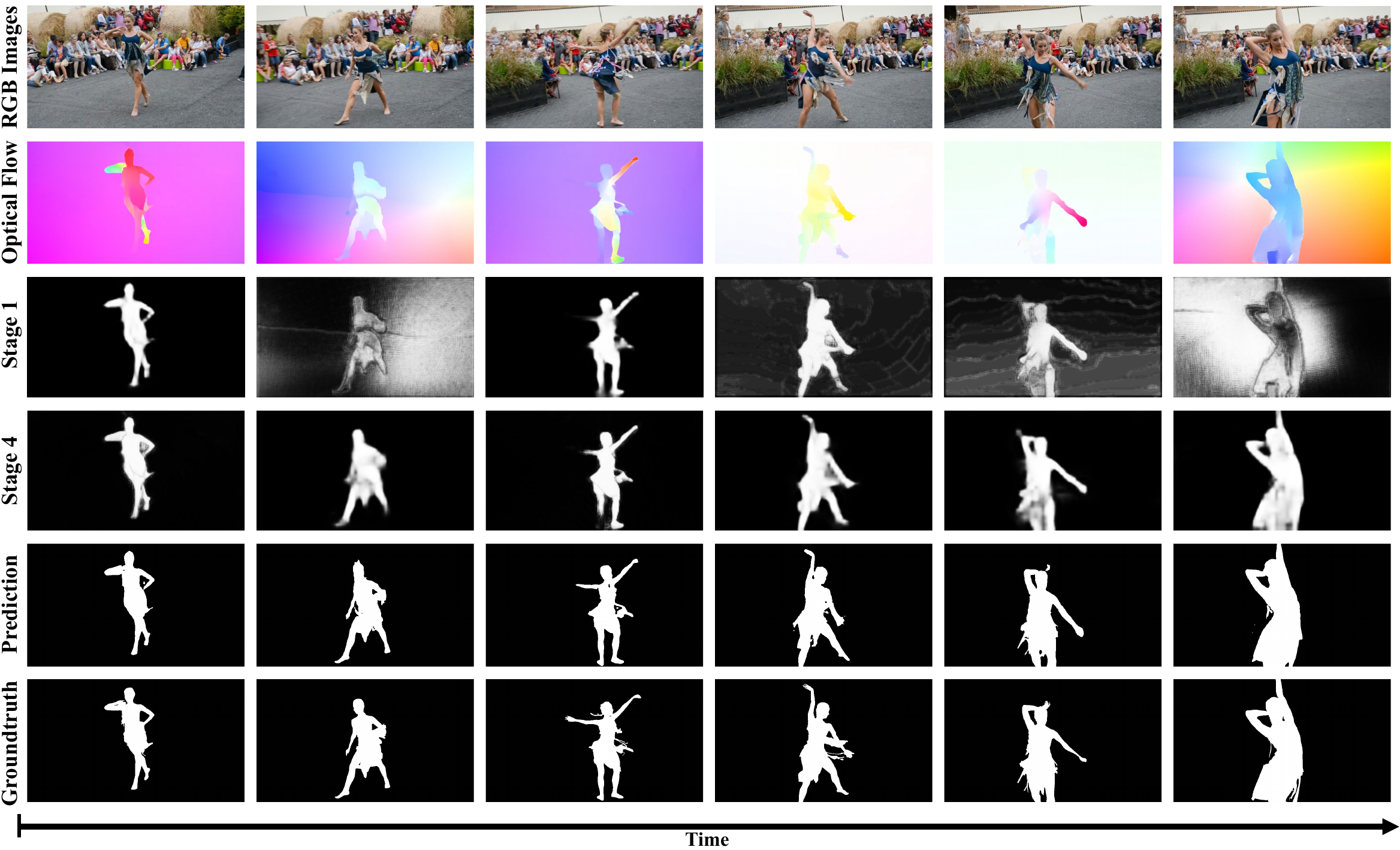}
  \caption{Visualization results of coarse mask in the first and fourth stage.}
  \label{fig:show_mask_ab}
\end{figure*}

\subsection{Experimental Setup}
In this study, we utilize three datasets for model training: DUTS~\cite{wang2017learning}, YouTube-VOS~\cite{xu2018youtube}, and DAVIS-16~\cite{perazzi2016benchmark}. We evaluate our SimulFlow on several popular unsupervised video object segmentation datasets: DAVIS-16~\cite{perazzi2016benchmark}, FBMS~\cite{ochs2013segmentation}, and YouTube-Objects~\cite{prest2012learning} to validate the effectiveness of SimulFlow. Following the standard settings, we adopt three evaluation metrics: region similarity  $\mathcal{J}$, boundary accuracy $\mathcal{F}$, and their average $\mathcal{G}$. For DAVIS-16, $\mathcal{J}$, $\mathcal{F}$, and $\mathcal{G}$ are used for evaluation, and for FBMS and YouTube-VOS, we only report $\mathcal{J}$.

\subsection{Implement Details}
\textbf{Architectures.} 
We design three kinds of structure with different parameters by varying the number of simulflow attention layers in each stage. 
The hyper-parameters of these variants are: (1) \textbf{SimulFlow-Small}: $N_{1}=2, N_{2}=2, N_{3}=2, N_{4}=2$; (2) \textbf{SimulFlow-Medium}: $N_{1}=3, N_{2}=3, N_{3}=6, N_{4}=3$; (1) \textbf{SimulFlow-Large}: $N_{1}=3, N_{2}=3, N_{3}=18, N_{4}=3$. We initialize the three models with MiT-b1, MiT-b2, and MiT-b3~\cite{xie2021segformer} respectively.

\textbf{Traininig.} 
The training sets includes DUTS, YouTube-VOS, and DAVIS-16. For the pretraining process on DUTS, we train model for 300K iterations and the learning rate is set as $6e^{-5}$ with a poly schedule. Then, we train model on YouTube-VOS for 200K iterations. The learning rate is set as $6e^{-5}$ with a poly schedule. Finally, the finetuning is conducted on DAVIS-16 for 5K iterations and the learning rate is fixed as $1e^{-6}$. During training, we utilize AdamW optimizer to optimize our model and implement data augmentation including random horizontal flipping, random resizing, and random cropping to $512 \times 512$. $\lambda$ is set as 0.1.

\subsection{State-of-the-art Comparison}
We compare the results of our SimulFlow with state-of-the-art unsupervised video object segmentation models on several popular benchmarks~\cite{perazzi2016benchmark,ochs2013segmentation,prest2012learning} to verify the effectiveness of our proposed model. Results are shown in Table~\ref{Table:results}. 

\textbf{DAVIS-16.}
DAVIS-16~\cite{perazzi2016benchmark} is one of the most popular UVOS datasets, which contains 30 training videos and 20 validation videos. Our SimulFlow outperforms all previous models by a significant margin with the highest speed and lowest parameters. Specifically, our SimulFlow-Small achieves 87.4\% $\mathcal{J} \& \mathcal{F}$ and reaches 63.7 FPS. Some methods such as AMC-Net~\cite{yang2021learning} and FSNet~\cite{ji2021full}, include post-processing techniques, which introduce heavy computational burden. 3DCSeg~\cite{mahadevan2020making} achieves competitive results, but the 3D convolutional layers require a large number of video training data. IMP~\cite{lee2022iteratively} depends on global information and cannot be implemented in an online manner. Without any post-processing techniques, our SimulFlow-Small achieves the balance between performance and speed and maintains online availability. Besides, our SimulFlow-Meduim and SimulFlow-Large scores 88.3\% and 89.1\% $\mathcal{J} \& \mathcal{F}$, respectively. The parameters of SimulFlow-Small are the lowest with just 13.7 MB for SimulFlow-Small, which can be easily employed on mobile devices. The SimulFlow attention allows simultaneous feature extraction and target identification is of significance to the high efficiency and effectiveness of SimulFlow.

\textbf{FBMS.} FBMS~\cite{ochs2013segmentation} is comprised of 59 videos with 29 training sequences and 30 test sequences. Our SimulFlow also achieves the best performances on FBMS. SimulFlow outperforms other models at least 0.5\% $\mathcal{J}$, which demonstrates that SimulFlow's robustness for handling videos with multiple.

\textbf{YouTube-Object.} YouTube-Object~\cite{prest2012learning} contains 126 videos over 20K frames and features 10 semantic categories, which is more challenging than DAVIS-16 and FBMS. We compare the mean performance of each category. SimulFlow also surpasses other models, which demonstrates the generalization ability of our model.

\begin{figure*}[htbp]
  \centering
  \includegraphics[width=\linewidth]{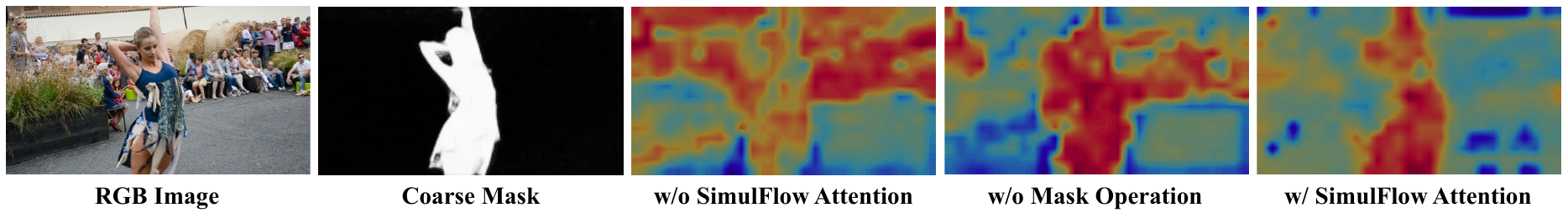}
  \caption{Visualization results of attention weights with different kind of attention operation.}
  \label{fig:show_attn}
\end{figure*}

\subsection{VSOD Results}
Video salient object detection (VSOD) aims to detect the salient object in a video and segment the salient object in each frame, which is similar to unsupervised video object segmentation. To verify the performance and generalization ability of our SimulFlow, we also conduct a quantitative comparison with state-of-the-art VSOD models on several VSOD datasets, including DAVIS-16~\cite{perazzi2016benchmark}, ViSal~\cite{li2013video}, and DAVSOD~\cite{fan2019shifting}. While SimulFlow is trained solely on unsupervised video object segmentation datasets, VSOD models incorporate extra VSOD datasets such as DAVSOD (except for DAVIS-16). We adopt three popular metrics to evaluate the performance: mean absolute error (MAE), structure measure $S_{\alpha} (\alpha=0.5)$, and max F-measure $F_{\beta}^{max}$ ($\beta^{2}=0.3$). Table \ref{tab:results_vsod} reports the results of our experiments, which demonstrate the considerable accuracy of SimulFlow on these benchmarks. The strong performance indicates the promising generalization ability of our SimulFlow on the VSOD task even without any specific training on VSOD datasets.

\subsection{Ablation Study}
\label{sec:ablation}

To verify the effectiveness of each component of SimulFlow, we perform an ablation study on DAVIS-16 validation set with SimulFlow-Small.
Results are shown in Table~\ref{tab:ablation}.

\textbf{Backbone.} The comparison of different backbone networks is presented in Table~\ref{tab:ablation_baseline}, where we analyze the trade-off between accuracy and speed. As the size of the backbone increases, the performance of SimulFlow improves, but with a corresponding increase in parameters and decrease in speed. Specifically, SimulFlow-Small, which adopts MiT-b1 as the backbone, achieves an accuracy of 87.4\% $\mathcal{J} \& \mathcal{F}$ with a speed of 63.7 FPS and only 13.7 MB parameters. SimulFlow-Medium, using MiT-b2 as backbone, achieves an accuracy of 88.3\% $\mathcal{J} \& \mathcal{F}$ with a speed of 36.5 FPS and 27.4 MB parameters. Finally, SimulFlow-Large, which employs MiT-b3 as the backbone, achieves the highest accuracy of 89.0\% $\mathcal{J} \& \mathcal{F}$ with a speed of 26.3 FPS and 47.2 MB parameters. Thanks to the proposed SimulFlow attention, SimulFlow strikes a good balance between performance and speed, making it suitable for practical applications.

\textbf{Cross Attention.} In SimulFlow attention, we just enable the cross attention from motion to image. We evaluated the impact of cross attention in SimulFlow attention by conducting several ablation experiments, as shown in Table~\ref{tab:ablation_cross}. In the first row, we first established a baseline scenario where we disabled all cross attention and mask operations, and only performed self attention in each stage. The resulting accuracy was 83.8\% $\mathcal{J} \& \mathcal{F}$. In the second row, cross attention from motion to image is enabled, which is the same as in simulflow attention. This approach resulted in a significant accuracy improvement to 85.6\% $\mathcal{J} \& \mathcal{F}$, but with a corresponding decrease in inference speed due to the extra computation introduced by cross attention.  In the third row, only cross attention from image to motion is enabled, resulting in a smaller improvement compared to the second row, indicating that the information flow from image to motion may bring about a negative influence. The fourth row allows for dual information flow, which facilitated performance, but at the expense of a significant drop in reasoning speed. To achieve a balance between speed and performance, an asymmetric design of SimulFlow attention was adopted, implementing only cross attention from motion to appearance.

\textbf{Mask Operation.} We conducted ablation experiments on the mask operation in SimulFlow attention to evaluate its effectiveness. Table~\ref{tab:ablation_mask} presents the results of different types of mask operation. In the second row, we applied a mask operation to the image feature, resulting in an accuracy improvement from 83.8\% $\mathcal{J} \& \mathcal{F}$ to 85.1\% $\mathcal{J} \& \mathcal{F}$, with a slight decrease in speed. This improvement demonstrates that mask operation succeeds in excluding noise from motion. In the third row, we tested a hard mask operation, which directly blocks out the background region instead of multiplying it by a low weight. The enhancement is lower than that in the second row, indicating that the hard mask may filter out some foreground areas due to prediction errors.  In the fourth row, we added soft mask attention to the motion feature, while in the fifth row, we added a mask to both image and motion features. Results indicate that adding mask operation to the motion feature may have a negative influence. 

We visualize the coarse mask at the first and fourth stage and the prediction in Figure~\ref{fig:show_mask_ab} to provide a clear understanding of how our SimulFlow model works and how it gradually refines its prediction to achieve accurate and robust segmentation. The initial coarse mask at the first stage captures the overall shape of the object but includes some noise due to the lack of detailed information. As the model progresses through each stage, the coarse mask becomes more refined and detailed, with fewer noise and more accurate object boundaries. At the final stage, the coarse mask closely resembles the groundtruth, demonstrating the effectiveness of our SimulFlow model in capturing object motion and appearance. We also visualize the image attention map of different kinds of operation in Figure~\ref{fig:show_attn} to verify the effectiveness of SimulFlow attention. The feature of the image is unaware of the target object without cross attention. Mask attention adds some constraint to the image feature, but there still exists too much noise in the background. Our SimulFlow attention successfully identifies salient objects with the help of motion feature and excludes motion noise.

\begin{figure*}[htbp]
  \centering
  \includegraphics[width=\linewidth]{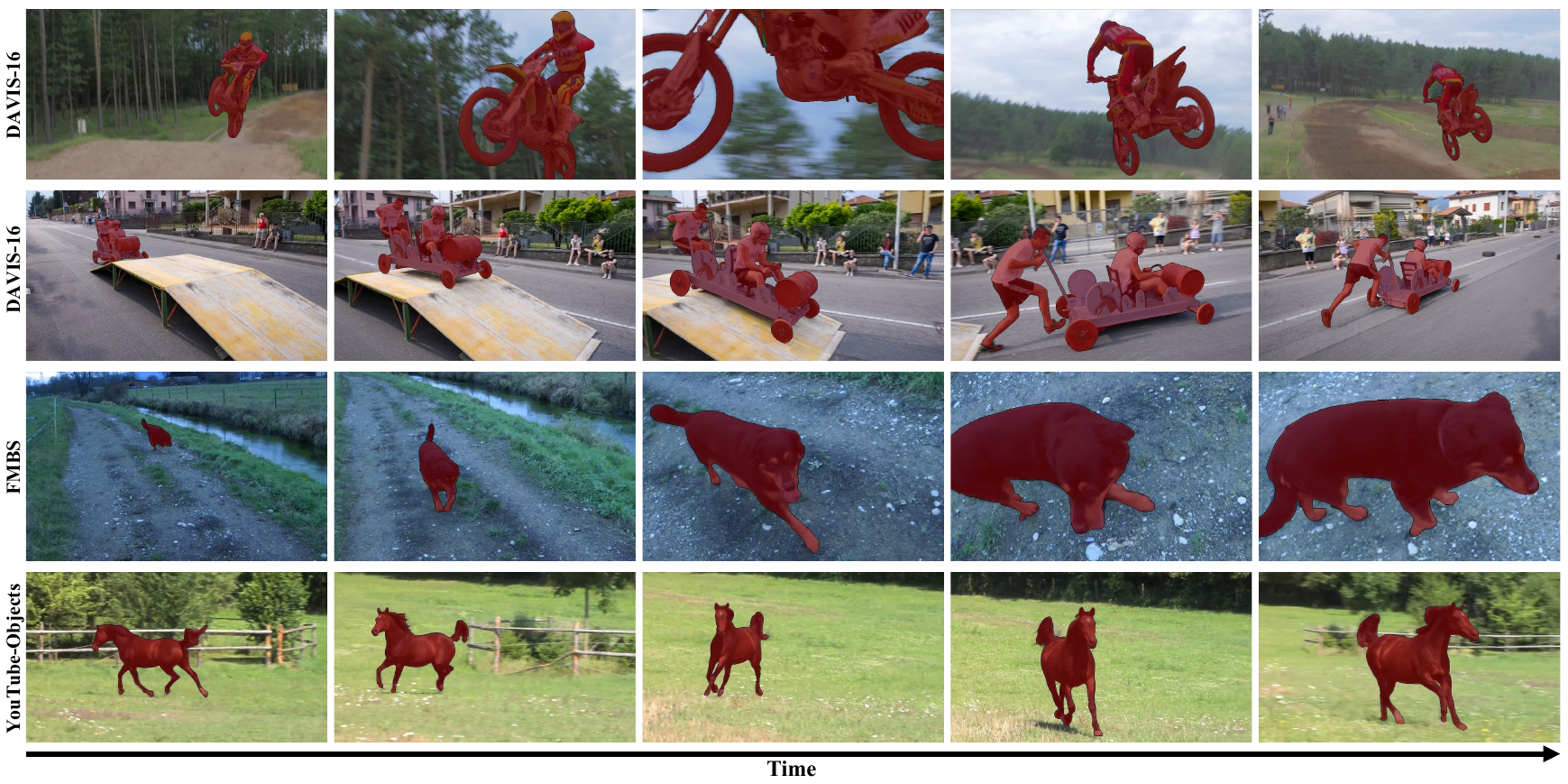}
  \caption{Qualitative results on DAVIS-16, FBMS, and YouTube-Object.}
  \label{fig:show}
\end{figure*}

\textbf{Stages of SimulFlow Attention.} To gain a deeper understanding of the impact of SimulFlow attention, we conducted experiments where we added SimulFlow attention to different stages, and compared the resulting performance in Table~\ref{tab:ablation_stage}. We observed that as the number of stages with SimulFlow attention increases, the overall performance of our model consistently improves. This finding highlights the importance of cross-model integration for achieving better results.

\textbf{Training Strategy.} Table~\ref{tab:ablation_train} presents the results of our ablation experiments to evaluate the impact of different training strategies on performance. In the first row, we train the model solely on DAVIS-16, which results in an unsatisfactory accuracy of 80.2\% $\mathcal{J} \& \mathcal{F}$ due to the limited number of training videos. To address the issue of data scarcity, we incorporate pretraining on either YouTube-VOS or DUTS, followed by fine-tuning on DAVIS-16, as shown in the second and third rows, respectively. Although pretraining helps improve performance, the model still suffers from overfitting on the small DAVIS-16. In the fourth row, we adopt the three-step training strategy, which yields the best performance and demonstrates the effectiveness of our training approach.

\subsection{Qualitative Results}
Figure~\ref{fig:show} exhibits some qualitative results of our SimulFlow model. We select four videos from DAVIS-16, FBMS, and YouTube-Object. Our model is robust to several challenges in unsupervised video object segmentation task, such as fast motion, scale variation, occlusion, and dynamic background.

\section{Conclusion}
In this work, we propose a simple and end-to-end unsupervised video object segmentation framework, SimulFlow, which drops the conventional two-stream pipeline and unifies the feature extraction and target identification. SimulFlow achieves a great balance between accuracy and speed. while also incorporating a SimulFlow attention mechanism to enable joint feature extraction and cross-modal integration. Experiments on several benchmarks demonstrate the effectiveness and generalization ability of our model, providing a promising solution for unsupervised video object segmentation that does not require a hand-designed fusion module. Our hope is that our work will inspire further research in this area, leading to more practical applications in real-world scenarios. 

\begin{acks}
This work was supported by National Natural Science Foundation of China (No.62072112), Scientific and Technological Innovation Action Plan of Shanghai Science and Technology Committee (No.22511102202), National Key R\&D Program of China (2020AAA0108301).
\end{acks}

\bibliographystyle{ACM-Reference-Format}
\bibliography{sample-base}


\end{document}